\documentclass[11pt]{article}
\usepackage{coling2020}
\usepackage[table]{xcolor}
\usepackage{algorithm}
\usepackage{algorithmic}
\usepackage{times}
\usepackage{tabularx}
\usepackage{graphicx}
\usepackage{url}
\usepackage{amsmath}
\usepackage{amssymb}
\usepackage{gb4e}
\usepackage{caption}
\usepackage{subcaption}
\usepackage{latexsym}

\colingfinalcopy 

\title{A Systematic Analysis of Morphological Content in BERT Models for Multiple Languages}

\author{Daniel Edmiston \\
  Department of Linguistics\\
  University of Chicago \\
  Chicago, IL 60615, USA \\ \\
  {\tt danedmiston@uchicago.edu}}

\date{}

\begin{document}
\maketitle
\begin{abstract}
	This work describes experiments which probe the hidden representations of several BERT-style models for morphological content. The goal is to examine the extent to which discrete linguistic structure, in the form of morphological features and feature values, presents itself in the vector representations and attention distributions of pre-trained language models for five European languages. The experiments contained herein show that (i) Transformer architectures largely partition their embedding space into convex sub-regions highly correlated with morphological feature value, (ii) the contextualized nature of transformer embeddings allows models to distinguish ambiguous morphological forms in many, but not all cases, and (iii) very specific attention head/layer combinations appear to hone in on subject-verb agreement.	
\end{abstract}

\section{Introduction}\label{Introduction} 

This work describes and reports on experiments designed to probe the contextualized word embeddings and attention distributions of several BERT-style Transformer models for discrete morphological structure. The experiments focus on testing the representations at the level of morphological feature and feature values, testing the extent to which these discrete structures are evident in the hidden-layer vectors and attention distributions produced by the models. By experimenting with five different languages, each somewhat different typologically, we hope to provide a general picture of how Transformer architectures model this aspect of morphological information.

To investigate the hidden representations of the models, we perform two types of experiment. For the first type, we perform a series of classification tasks on hidden-layer vector representations, attempting to predict the morphological feature values of contextualized representations. The second type of experiment consists of a task which probes self-attention distributions for what linguists call an agree relationship, comparing the proportions of a sentence's attention distributions allocated to words which agree for some morphological feature value.

By focusing on morphological information, this work complements much recent work devoted to probing the hidden representations of BERT-style models for syntactic and semantic information. We contend that analysis at the level of morphological feature is particularly useful for evaluating linguistic information within embeddings for three reasons. First, morphological features represent a tangible aspect of meaning for which it is relatively simple to obtain large amounts of quality gold-standard annotation in many languages. Second, by evaluating at the level of morphological feature, experiments are less susceptible to models using heuristics to learn surface patterns \cite{McCoyEtAl2019}, testing instead whether they've generalized to the underlying cause of those patterns. Finally, certain morphological features contain aspects of meaning which are vitally important for real-world tasks. For instance, the gender feature is inextricably linked to coreference resolution in many languages, and the morphological feature of mood contains discourse information which is necessary for natural language understanding, for example distinguishing between commands (\textit{imperative} mood) and statements (\textit{indicative} mood).

The contributions of this paper are therefore the following: we show that (i) BERT-style architectures are capable of encoding morphological information in their hidden vector representations at the featural level, and do so by dividing the embedding space into convex (i.e. linearly separable) sub-regions, (ii) that the contextualized nature of embeddings aids models' ability to disambiguate morphologically ambiguous forms, but doesn't solve the problem, and (iii) by introducing a score based on Pearson's $\chi^{2}$ test for investigating attention distributions, we show that localized regions in the layer/attention-head space reflect subject-verb agreement.

\section{Related Work}\label{Related Work}

With the recent success of Transformer-style architectures \cite{VaswaniEtAl2017} like BERT \cite{DevlinEtAl2018} on many natural language processing tasks, a considerable amount of research has gone into investigating the inner workings of these models, a research program sometimes dubbed ``BERTology'' \cite{RogersEtAl2020}. Among this literature, work has focused on syntactic aspects \cite{HewittManning2019,CoenenEtAl2019,KimEtAl2020}, including subject-verb agreement \cite{Goldberg2019}, and also various semantic aspects such as semantic role and model predictions' correlation with human judgment \cite{Ettinger2020}. 

One particular method of probing the information in these large architectures is to perform different tasks at different layers of the model, seeking to identify where different types of linguistic information may reside \cite{TenneyEtAl2019a}. It has generally been shown that more local, shallow information is reflected in lower layers, and richer, more abstract information is reflected in higher layers \cite{PetersEtAl2018b}. Not only have layers been shown to be specialized for content, \newcite{ClarkEtAl2019} further showed the diffusion of linguistic knowledge through such models by demonstrating that BERT's different attention heads learn to focus on different aspects of linguistic meaning.

In addition to the growing literature on BERT-style models, work on evaluating continuous embedding models for morphological information goes back some years. Particularly, \newcite{BelinkovEtAl2017} trained classifiers on word representations extracted from models trained for machine translation to assess what these models learn about morphology. For work investigating models for agree-phenomena, \newcite{LinzenEtAl2016} showed that LSTM architectures \cite{HochreiterSchmidhuber1997} successfully model subject-verb agreement in many instances (see also \newcite{GiulianelliEtAl2018}), and \newcite{RavfogelEtAl2018} put forth the objective of modelling agreement in Basque as a potential baseline for future work.. Finally, for work most similar to ours in investigating morphological information at the featural level, \newcite{BasiratTang2018} train classifiers to distringuish nominal features in Swedish, and \newcite{Kohn2015} does the same for more varied features and multiple languages.

This work finds its place in systematically addressing the question of morphological featural information in the hidden vector representations and attention heads of BERT models for multiple languages.

\section{Methodology}\label{Methodology}

\subsection{Considered Languages and their Morphology}\label{Morphology}

This study investigates five languages of the Indo-European language family: English, French, German, Russian, and Spanish, each of which inflects for some set of morphological features. Specifically, we investigate the morphological features of Case, Gender, Mood, Number, Person, Tense, and Verb Form (which is related to what is traditionally called Finiteness). For each language, we investigate a BERT-base model \cite{DevlinEtAl2018} pre-trained on a large corpus for that language.\footnote{See \newcite{DevlinEtAl2018} for details on the English model, \newcite{MartinEtAl2019} for the French model, and \newcite{KuratovArkhipov2019} for the Russian model.} The languages, models, features, and each feature's values are ogranized in Table \ref{LanguagesAndMorphology}.\footnote{Abbreviations used in Table \ref{LanguagesAndMorphology}: Nom=Nominative, Acc=Accusative, Dat=Dative, Gen=Genitive, Loc=Locative, Ins=Instrumental, Masc=Masculine, Fem=Feminine, Neut=Neuter, Ind=Indicative, Imp=Imperative, Sub=Subjunctive, Cnd=Conditional, Sing=Singular, Plur=Plural, Pres=Present, Impr=Imperfect, Fut=Future, Fin=Finite, Inf=Infinitive, Ger=Gerund, Part=Participle, Conv=Converb.}

\begin{table}
	\begin{center}
		\scalebox{.75}{
			\begin{tabular}{c|c|c|c|c|c}
				\hline 
				Feature $\backslash$ Language & English & French & German & Russian & Spanish \\
				\hline\hline 
				Case & \cellcolor{lightgray} & \cellcolor{lightgray} & $\{Nom, Acc, Dat,$ & $\{Nom, Acc, Dat,$ & \cellcolor{lightgray} \\
				
				& \cellcolor{lightgray} & \cellcolor{lightgray} & $Gen\} $ & $ Gen, Loc, Ins \}$ & \cellcolor{lightgray} \\ 
				
				\hline 
				
				Gender & \cellcolor{lightgray} & $\{ Masc, Fem \}$ & $\{ Masc, Fem, Neut \}$ & $\{Masc, Fem, Neut\}$ & $\{Masc, Fem\}$ \\
				
				\hline 
				
				Mood & $\{Ind, Imp\}$ & $\{ Ind, Sub, Cnd$ & $\{ Ind, Sub, Imp \}$ & $\{ Ind, Cnd, Imp \}$ & $\{ Ind, Sub, Cnd,$ \\
				
				& & $ Imp \}$ & & & $ Imp \}$ \\
				
				\hline 
				
				Number & $\{Sing, Plur\}$ & $\{Sing, Plur\}$ & $\{Sing, Plur\}$ & $\{Sing, Plur\}$ & $\{Sing, Plur\}$ \\
				
				\hline 
				
				Person & $\{ 1, 2, 3 \}$ & $\{ 1, 2, 3 \}$ & $\{ 1, 2, 3 \}$ & $\{ 1, 2, 3 \}$ & $\{ 1, 2, 3 \}$ \\
				
				\hline 
				
				Tense & $\{ Past, Pres \}$ & $\{ Past, Pres, Impr,  $ & $\{ Past, Pres \}$ & $ \{Past, Pres, Fut\} $ & $\{ Past, Pres, Impr,  $  \\
				
				& & $ Fut \}$ & & & $ Fut \}$ \\
				
				\hline 
				
				Verb Form &  $\{ Fin, Inf, Ger, $ & $\{Fin, Inf, Part\}$ & $\{Fin, Inf, Part\}$ & $\{ Fin, Inf, Part$ & $ \{ Fin, Inf, Part, $ \\
				
				& $ Part \}$ & & & $ Conv \}$ & $ Ger \}$ \\
				
				\hline\hline

				Model & Base-Cased & CamemBERT & DBMDZ & RuBERT & BETO \\

				\hline 
				
		\end{tabular}}
		\caption{Languages and associated features/feature values, along with models used. All models are BERT-base models, with 12 hidden layers, 12 attention heads, and 768 dimenional vectors. All models available at \url{https://huggingface.co/models}.}
		\label{LanguagesAndMorphology}
	\end{center}
	
\end{table}

\subsection{Experiments}\label{Experiments}

\subsubsection{Experiment 1: Classifying by Value}\label{Experiment1}

As BERT-base models are a 12-layer transformer architecture, for each word in a sentence the model produces 13 vectors, including the input vector, in 768 dimensions.\footnote{Strictly speaking, the model produces vectors for each token in a sentence. We derive word embeddings by taking the average of each word's constituent token embeddings.} To test the amount of morphological information in a model's hidden vectors, for each layer we perform $n$-way classification tasks for each feature, where $n$ is the number of feature values that the relevant feature can take (e.g. $n=3$ for Mood in German, values being Indicative, Imperative, and Subjunctive).

The classification tasks are $k$-means clustering, a linear classifier, and a non-linear classifier, amounting to a 3-layer neural network with ReLU-activations. For the $k$-means task, the best score is taken from amongst ten runs with different centroid seeds. (Weighted) F1 scores are calculated for each experiment.

\subsubsection{Experiment 2: Subject-Verb Agreement in Attention}\label{Experiment2}

The second experiment tests whether what linguists call agreement presents itself in the attention coefficients produced by BERT-style models when embedding sentences. Agreement is a syntactic phenomenon in which two syntactic constituents in a certain relationship show agreement for some morphological feature. An example from English showing agreement for Number is below.

\begin{exe}
	\ex The \textbf{men} \textbf{were} tired after a hard day's work.
\end{exe}

In more morphologically rich languages like French, the agree relation can easily encompass every word in a sentence. 

\begin{exe}
	\ex \gll Les grands gar\c cons sont tous all\'es \\ 
	the.\textsc{plur} tall.\textsc{plur} boy.\textsc{plur} are.\textsc{plur} all.\textsc{plur} left.\textsc{plur} \\
	
	\glt The tall boys all left
\end{exe}

The question is how to investigate the output of an attention head for some layer, call it an attention-matrix, for awareness of this agree relation. Considering that an attention-matrix for a sentence $s=w_{1},...,w_{n}$ can be interpreted as a sequence of $n$ probability distributions over $n$ words, we can intuit that an attention-matrix reflects the agree relation if the attention distributions for the words inside the agree relation place a disproportionate amount of probability mass on the other words in the agree relation, and those words outside the agree relation do not.

Then given such an example sentence showing the relevant agree relation, we can quantify the extent to which some attention-matrix from a BERT model reflects this relation using a method based on the $\chi^{2}$-test in the following way. Given sentence $s=w_{1},...,w_{n}$ , partition $s$ into an $Agree$-$Set=\{w_{i} \text{ }\vert\text{ }w_{i} \text{ participates in agree} \}$ and an $Out$-$Set=\{w_{i} \text{ }\vert\text{ }w_{i} \text{ does not participate in  agree} \}$, thusly partitioning the matrix of attention distributions into Agree-distributions and Out-distributions.

%

For each attention distribution in the Agree-distributions, calculate the Pearson's $\chi^{2}$ score, where the two possible outcomes are $Agree$ and $Out$;\footnote{That is, for each distribution calculate the Pearson $\chi^{2}$-score for the probability mass allotted to words in the Agree-set vs. the probability mass allotted to words in the Out-set.} calculate the average $\chi^{2}$-score for the distributions in the Agree-set. Repeat the process for the distributions in the Out-set. Per our intuition that an attention-matrix reflects the agree relation insofar as it allots a disproportionate amount of probability mass to words which participate in agree for distributions of words in agree, and does not do so otherwise, then a high Agree-score, with a relatively low Out-score, means an attention-matrix focuses probability mass disproportionately between words which participate in the agree relation. The point of considering the Agree-score against the Out-score is to account for the contingency in which words participating in agree are particularly salient for reasons other than agree.\footnote{For all attention-matrices, the diagonals were set to 0 and the probability distributions renormalized to sum to 1. This was done to account for the fact that heads have a tendancy to focus a large amount of mass on the diagonal, which would skew towards higher Agree-scores for words in the Agree-set, and towards lower scores for words in the Out-set.}

\subsection{Data Sets}\label{DataSets}

All data used for the experiments were collected from a collection of Universal Dependency (UD) Treebanks \cite{NivreEtAl2016} and UD-compatible lexicons \cite{Sagot2018}. See Appendix \ref{Treebanks} for specifics on the treebanks used. For the classification tasks described above, for each language-feature combination examples of the following form were extracted, (\textit{word}, \textit{sentence}, \textit{value}), the task being to embed the word in the context of its sentence in order to predict its value. For each such classification dataset, 750 examples for each value were sampled,\footnote{The only exceptions to this was were the Mood feature in French and Spanish, for which there was insufficient data to extract 750 examples of the imperative. As such, the French Mood dataset consisted only of 249 examples for each value, and the Spanish mood dataset only 381 examples for each value.} with .85/.15 train-test splits for the supervised tasks.

An important aspect of morphology in Indo-European languages like those chosen for this study is both the number of values a feature can take---call it feature-length---and also the amount of syncretism they display \cite{BaermanEtAl2005}, that is, how likely the forms of a language are to be ambiguous for value. For instance, the definite determiner `\textit{der}' in German can be nominative, dative, or genitive in Case value depending on context. The amount of such ambiguous forms is an important consideration when classifying. 

\begin{table} 
	\centering
	\begin{tabular}{c|c|c|c|c|c}
		\hline 
		Confound$\backslash$Language  & English & French & German & Russian & Spanish \\
		\hline 
		Pct. of ambiguous forms & 18.1\% & 10.3\% & 26.0\% & 14.1\% & 6.1\% \\
		Avg. feature-length & 2.6 & 3.0 & 2.86 & 3.43 & 3.17 \\
		\hline 
	\end{tabular}
	\caption{Percentage of ambiguous examples across all features for language, as well as average feature length for relevant features.}
	\label{ComplexityStats}
\end{table}

Table \ref{ComplexityStats} describes the statistics of the datasets for each language with regard to the percentage of ambiguous examples and average feature length. Intuitively, one would expect performance to be negatively correlated with feature-length and confounds such as ambiguity. This expectation is borne out in Section \ref{EffectsOfComplexity}.

For the agree-related tasks discussed in Section \ref{Experiment2}, examples of subject-verb agreement were collected for English, French, and German. For English, only noun-verb pairs agreeing for the Number feature and marked with the \textit{nsubj} dependency between noun and verb were chosen, and only when one such dependency was present in the sentence. For French and German, which each show richer agreement phenomena than English, examples of subject-verb agreement were chosen such that the subject was of the form Det-Adj-Noun (or Det-Noun-Adj), and all words agreed for the number feature, again with the \textit{nsubj} dependency and only one-example-per-sentence criteria holding. For the English and German treebanks, 2,000 examples were extracted, and 1,521 examples were extracted from the French treebanks.

\section{Results}

\subsection{Results on Classification Tasks}

\subsubsection{Results by Feature}

Table \ref{ClassificationResultsByFeature} displays (weighted) F1 scores for each language for each relevant feature and for each classification task. The scores in the table are averaged over all layers. Random baselines for this task are found in Table \ref{ClassificationResultsByFeatureRandom} in Appendix \ref{FullResults}.\footnote{Code to reproduce results can be found at \url{https://github.com/danedmiston/morphology_classifiers}.}

\begin{table*}[h!]
	\small
	\begin{center}
		\begin{tabularx}{\linewidth}{X|X|X|X|X|X||X}
			\hline 
			Language  & English & French & German & Russian & Spanish & Average \\
			\hline
		\end{tabularx}
		\begin{tabularx}{\linewidth}{XXX|XXX|XXX|XXX|XXX|XXX||XXX}
			Feature$\backslash$Task &&& KM & Lin & NN & KM & Lin & NN & KM & Lin & NN & KM & Lin & NN & KM & Lin & NN & KM & Lin & NN \\
			\hline 
			
			Case     &&& \cellcolor{lightgray}        &  \cellcolor{lightgray}       &  \cellcolor{lightgray}       &      \cellcolor{lightgray}  &   \cellcolor{lightgray}     &  \cellcolor{lightgray}      & 0.25   & 0.87   & 0.88   & 0.15    & 0.84    & 0.86    &   \cellcolor{lightgray}      &  \cellcolor{lightgray}       &   \cellcolor{lightgray}      & 0.2     & 0.86    & 0.87    \\
			
			Gender   &&&  \cellcolor{lightgray}       &   \cellcolor{lightgray}      &   \cellcolor{lightgray}      & 0.5    & 0.96   & 0.96   & 0.32   & 0.9    & 0.91   & 0.28    & 0.86    & 0.87    & \textbf{0.48}    & 0.97    & 0.97    & 0.39    & 0.92    & 0.93    \\
			
			Mood     &&& 0.6     & 0.98    & 0.98    & 0.3    & 0.98   & 0.96   & 0.41   & 0.91   & 0.85   & 0.28    & \textbf{0.99}    & \textbf{0.99}    & 0.26    & 0.9     & 0.88    & 0.37    & 0.95    & 0.93    \\
			
			Number   &&& 0.46    & 0.97    & 0.97    & 0.49   & 0.97   & 0.97   & 0.48   & 0.93   & 0.92   & \textbf{0.48}    & 0.92    & 0.93    & 0.46    & \textbf{0.99}    & \textbf{0.99}    & \textbf{0.47}    & 0.96    & 0.95    \\
			
			Person   &&& \textcolor{red}{0.32}    & 0.96    & 0.95    & \textcolor{red}{0.33}   & 0.97   & 0.97   & 0.39   & 0.96   & 0.95   & \textcolor{red}{0.25}    & \textbf{0.99}    & 0.96    & 0.31    & 0.93    & 0.93    & 0.32    & 0.96    & 0.95    \\
			
			Tense    &&& \textbf{0.64}    & \textbf{1.0}     & \textbf{0.99}    & 0.25   & 0.99   & \textbf{0.99}   & \textbf{0.55}   & \textbf{0.97}   & \textbf{0.97}   & 0.34    & 0.97    & 0.96    & 0.21    & 0.98    & 0.96    & 0.4     & \textbf{0.98}    & \textbf{0.97}    \\
			
			VerbForm &&& \textcolor{red}{0.2}     & 0.88    & 0.87    & 0.39   & \textbf{1.0}    & 0.96   & 0.33   & \textbf{0.97}   & 0.94   & \textcolor{red}{0.2}     & \textbf{0.99}    & \textbf{0.99}    & 0.33    & 0.98    & 0.98    & 0.29    & 0.96    & 0.95    \\
			
			\hline 
			\hline 
			
			Average  &&& 0.44    & 0.96    & 0.95    & 0.38   & 0.98   & 0.97   & 0.39   & 0.93   & 0.92   & 0.28    & 0.94    & 0.94    & 0.34    & 0.96    & 0.95    & \cellcolor{lightgray}    & \cellcolor{lightgray}    & \cellcolor{lightgray}  \\
			\hline 
		\end{tabularx}
		
	\end{center}
	\caption{Weighted F1 Scores for each language and feature; scores averaged across all layers. KM=K-Means clustering, Lin=Linear classifer, NN=3-layer Neural Network with ReLU activations. Bold reflects highest score in each column. Red indicates score $\leq$ random baseline (see Table \ref{ClassificationResultsByFeatureRandom}).}
	
	\label{ClassificationResultsByFeature}
\end{table*}

The results indicate that each language's model reflects a great deal of morphological information at the featural level. However, it appears that supervision aids tremendously in extracting this information; the K-Means clustering scores in most cases fall very near the random baseline scores for the same task, and in some cases below. Meanwhile, in the case of the linear and non-linear classifiers, average performance of the pre-trained models significantly surpasses the random baselines.

Furthermore, the fact that linear classifiers routinely returned F1 scores above 0.9, and in two cases perfect scores, strongly suggests that pre-trained models are partitioning their embedding spaces into convex regions correlated with morphological feature value; the additional power of the non-linear neural network model did not significantly improve performance in most cases. The results further suggest that certain morphological features may be better captured than others. In the linear case, when compared to random baselines the two best performing features on average were VerbForm and Case; Mood and Person were the two worst. In terms of overall performance, the Tense feature was classified the most faithfully, and the Case feature the least. Here, the percentage of ambiguous forms likely played a role; see Section \ref{EffectsOfComplexity}. Further discussion follows in Section \ref{Discussion}.

\subsubsection{Results by Layer}

Above, we averaged over the layers to get a sense of how well these models were reflecting different morphological features in their vector representations. This section presents results where the scores are averaged over the features and presented for each layer. 

\begin{figure}[h]
	\centering
	\begin{subfigure}{.5\textwidth}
		\centering
		\includegraphics[scale=.33]{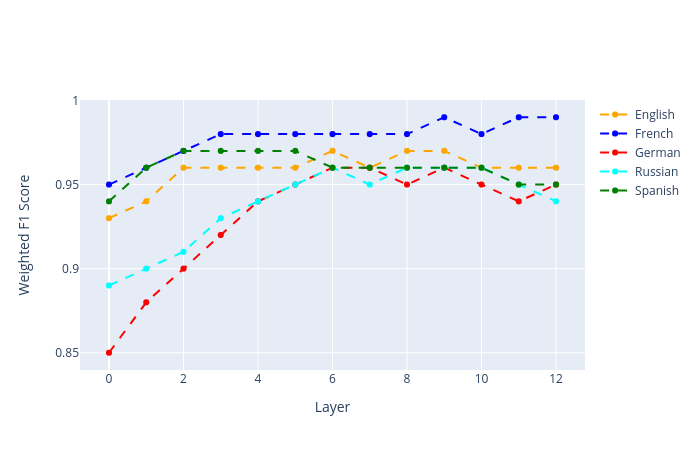}
	\end{subfigure}%
	\begin{subfigure}{.5\textwidth}
		\centering
		\includegraphics[scale=.33]{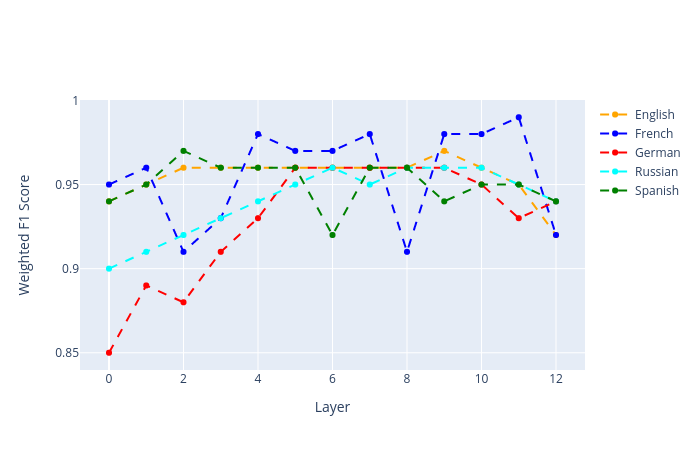}
	\end{subfigure}
	\caption{Layer-wise weighted F1 scores for linear (left) and neural network (right) classifiers averaged over all features.}
	\label{Layers}
\end{figure}

A visual inspection of the results from the linear classifer in Figure \ref{Layers} suggests that morphological information is best captured in the middle-late layers in German and Russian, with solid performance throughout for English, French, and Spanish. Speculation as to the reason for this can be found in Section \ref{Discussion}. The results of the non-linear classifier are more varied, but roughly coincide with the results of the linear classifier. See Appendix \ref{FullResults} for a full reporting on layer-wise scores (Table \ref{ClassificationResultsByLayer}), as well as random baseline scores for the same task (Table \ref{ClassificationResultsByLayerRandom}).

\subsubsection{Effects of Complexity}\label{EffectsOfComplexity}

As mentioned in Section \ref{DataSets}, the languages involved not only display ambiguous morphological forms, presumably making classification more difficult, certain features in certain languages also present the challenge of having a large number of possible values, as many as six in the case of the Russian Case feature. This section presents results which suggest that these factors do indeed make classification more difficult, showing that in most cases an increase in the percentage of ambiguous forms is negatively correlated with classification performance, as is an increase in the number of values a feature may take.

\begin{table*} 
	\centering 
	\begin{tabularx}{\linewidth}{X|X|X|X|X|X}
		\hline 
		Language & English & French & German & Russian & Spanish \\
		\hline 
	\end{tabularx}
	\begin{tabularx}{\linewidth}{XX|XX|XX|XX|XX|XX}
		
		\small Correlation$\backslash$Test  & & \small Perc. & \small \#-Vals. &\small Perc. & \small \#-Vals. &\small Perc. &\small \#-Vals. &\small Perc. &\small \#-Vals. &\small Perc. &\small \#-Vals. \\
		\hline 
		\small Spearman & & -0.80 & \cellcolor{blue!30}-0.89 & -0.68 & 0.6 & \cellcolor{blue!15}-0.75 & -0.54 & \cellcolor{blue!30}-0.89 & -0.11 & -0.09 & -0.25 \\
		
		\small Pearson & & \cellcolor{blue!15}-0.86 & \cellcolor{blue!30}-0.93 & \cellcolor{blue!15}-0.79 & 0.58 & \cellcolor{blue!30}-0.84 & -0.62 & \cellcolor{blue!30}-0.87 & -0.45 & -0.06 & -0.33 \\
		\hline  
	\end{tabularx}
	\caption{Spearman and Pearson correlation scores between performance on feature (as measured by the linear classifier in Table \ref{ClassificationResultsByFeature}) and percentage of ambiguous entries/number of possible values for feature. Light blue indicates statistically significant with $p$-value below 0.1, Dark blue indicates statistically significant with $p$-value below 0.05.}
	\label{AmbiguityCorrelation}
\end{table*}

The results in Table \ref{AmbiguityCorrelation} show that ambiguous forms predictably make classification more difficult in all cases, though with the situation being somewhat less pronounced in Spanish, in which there is very little morphological ambiguity. Likewise, as the number of possible values a feature may take increases, performance also suffers in all cases except French, which was the strongest performing language and performed well on all tasks. 

While the percentage of ambiguity for a feature is negatively correlated with weighted F1 performance, the depth of BERT-style models does go some way to alleviating this problem, as classification shows a general upward trend through the layers, peaking by the middle layers. 

\begin{figure}[h]
	\centering
	\begin{subfigure}{.5\textwidth}
		\centering
		\includegraphics[scale=.33]{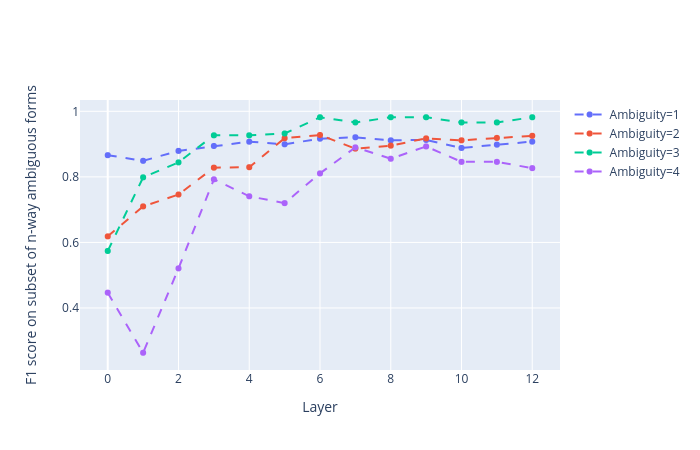}
		\label{fig:sub1}
	\end{subfigure}%
	\begin{subfigure}{.5\textwidth}
		\centering
		\includegraphics[scale=.33]{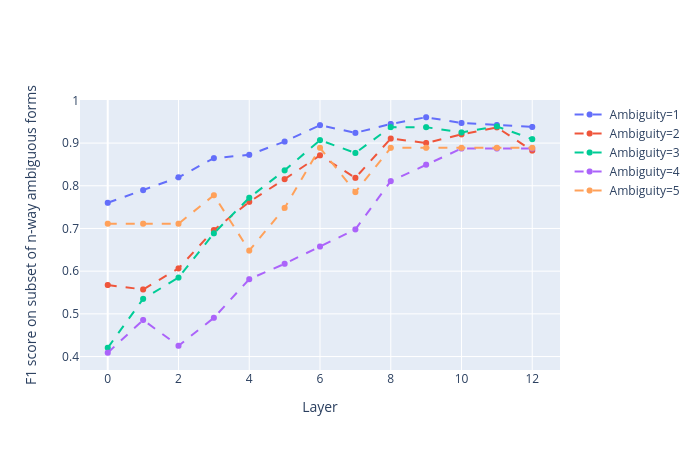}
		\label{fig:sub2}
	\end{subfigure}
	\caption{Performance per-layer on Case feature for linear classifier for German (left) and Russian (right). Both layers show strongest performance in middle layers.}
	\label{GermanRussianCase}
\end{figure}

Figure \ref{GermanRussianCase} shows the per-layer performance of German and Russian on $n$-way ambiguous subsets for the Case-feature, with forms in German being up to 4-way ambiguous and in Russian up to 5-way ambiguous. In all cases (except where ambiguity=1 in German, meaning the word in unambiguous) there is a pronounced upward trend in performance through the layers, suggesting that BERT is able to make use of context to disambiguate forms for morphological feature. However, it is worth keeping in mind that amount of ambiguity is still negatively correlated with performance, suggesting that BERT is far from human-like performance, in spite of its contextualized nature. 

\subsection{Results on Agree Task}

As described in Section \ref{Experiment2}, an attention-matrix's reflection of the agree relation can be relayed by two quantities: the Agree-score and the Out-score. Recalling the definitions of the Agree-score and Out-score as the average $\chi^{2}$-scores for distributions in the agree set and the non-agree set respectively, a high ratio of agree-score to out-score signifies that an attention-matrix focuses a disproportionately high amount of probability mass between words which agree, while words which don't participate in agree do not. 

\begin{table}[]
	\scalebox{.7}{
		\begin{tabular}{c|cccccccccccc}
			\hline 
			& Head=1 & Head=2 & Head=3 & Head=4 & Head=5 & Head=6 & Head=7 & Head=8 & Head=9 & Head=10 & Head=11 & Head=12 \\
			\hline 
			Layer=1  & 0.02   & 0.03   & 0.01   & 0.21   & \cellcolor{blue!15}3.32   & 0.02   & 0.02   & 0.01   & 0.02   & 2.52    & 0.03    & 0.01    \\
			Layer=2  & 0.04   & 0.47   & 1.11   & 0.1    & 0.04   & 0.03   & 0.03   & 0.03   & 0.05   & 1.93    & 0.5     & 0.02    \\
			Layer=3  & 0.09   & \cellcolor{blue!30}4.03   & 0.02   & 0.03   & 0.03   & 0.04   & 0.39   & 0.84   & 0.03   & 0.03    & 0.37    & 0.03    \\
			Layer=4  & 0.05   & 0.25   & 0.03   & 0.03   & 0.03   & 0.04   & 0.04   & 0.05   & 0.04   & 0.32    & 0.06    & 0.03    \\
			Layer=5  & 0.03   & 0.04   & 0.02   & \cellcolor{blue!15}3.59   & 0.04   & 0.03   & 0.04   & 0.03   & 0.04   & 0.03    & 0.26    & 0.19    \\
			Layer=6  & 0.62   & 0.03   & 0.03   & 0.04   & 0.36   & 0.05   & 0.06   & 0.03   & 0.03   & 0.03    & 0.03    & 0.57    \\
			Layer=7  & 0.04   & 0.12   & 0.04   & 0.02   & 0.04   & 0.03   & 0.04   & 0.03   & 0.04   & 0.03    & 0.39    & 0.03    \\
			Layer=8  & 0.3    & 0.03   & 0.05   & 0.1    & 0.04   & 0.03   & 0.16   & 0.04   & 0.03   & 0.06    & 0.03    & 0.08    \\
			Layer=9  & 0.03   & 0.03   & 0.92   & 0.05   & 0.04   & 0.08   & 1.21   & 0.05   & 0.04   & 0.02    & 0.08    & 0.04    \\
			Layer=10 & 0.09   & 0.11   & 0.99   & 0.04   & 0.02   & 0.05   & 0.04   & 0.04   & 0.03   & 0.07    & 0.03    & 0.08    \\
			Layer=11 & 0.04   & 0.07   & 0.04   & 0.03   & 0.04   & 0.61   & 0.03   & 0.04   & 0.04   & 0.02    & 0.19    & 0.03    \\
			Layer=12 & 0.03   & 0.03   & 0.79   & 0.02   & 0.04   & 0.12   & 0.04   & 0.03   & 0.03   & 0.03    & 0.03    & 0.04  \\
			\hline 
	\end{tabular}}
	\caption{Average agree-score (average $\chi^{2}$-score over distributions in agree-set) over French Agree dataset for each head/layer combination. Light blue shading denotes value exceeds $\chi^{2}$-score required for $p$-value$<0.1$ for one degree of freedom (2.706).}
	\label{FrenchAgree}	
\end{table}

Taking French as an initial example, Table \ref{FrenchAgree} shows that a small number of head-layer combinations are focusing a large (and statistically significant) amount of attention on the agreeing set, while most others treat the agreeing set as would be expected by chance. The results in Tables \ref{EnglishAgree} and \ref{GermanAgree} likewise show that the agree information is concentrated in few head-layer combinations in English and German, though the information is somewhat more diffuse and the scores higher than in French. 

In all languages there are combinations which show an average Agree-score (i.e. average $\chi^{2}$ score on agree-set distributions) which is statistically significant with a $p$-value$<.05$, and others which show an average Agree-score with significance $p$-value$<0.1$. Meanwhile, most head-layer combinations remain close to 0. In no language is there a significant average Out-score (full results are listed in Appendix \ref{FullResults}). The relatively high scores for Agree-set vs. Out-set, and the fact that the high Agree-scores are localized to a small number of head-layer combinations, suggests that certain head-layer combinations are in fact honing in on the agree relation in discriminating fashion; i.e. BERT-style pre-trained language models appear sensitive to subject-verb agreement. 

\begin{figure}[h]
	\centering
	\begin{subfigure}{.33\textwidth}
		\centering
		\includegraphics[scale=.33]{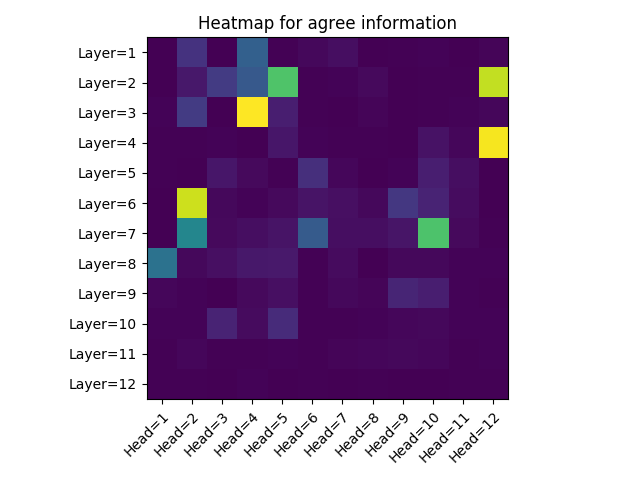}
		\label{EnglishAgreeMap}
	\end{subfigure}%
	\begin{subfigure}{.33\textwidth}
		\centering
		\includegraphics[scale=.33]{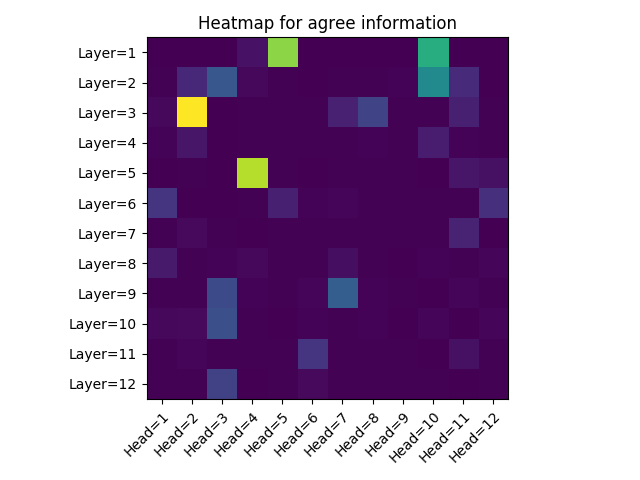}
		\label{FrenchAgreeHeatmap}
	\end{subfigure}
	\begin{subfigure}{.33\textwidth}
		\centering
		\includegraphics[scale=.33]{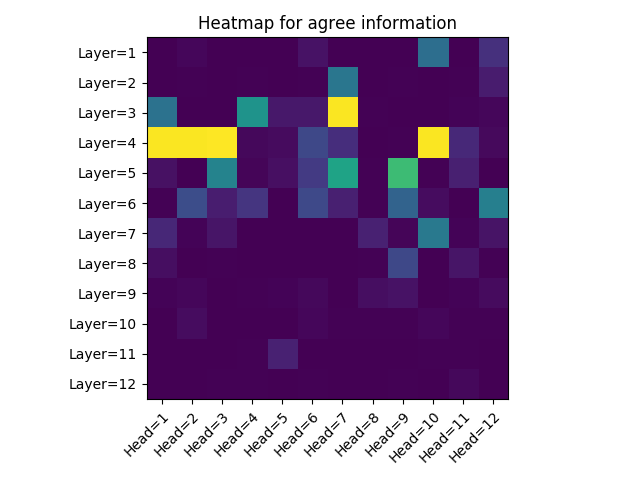}
		\label{GermanAgreeHeatmap}
	\end{subfigure}
	\caption{Heatmaps for attention head-layer combinations for average agree-score for English (left), French (center), and German (right); bright spots indicate high agree-score.}
	\label{Heatmaps}
\end{figure}

Figure \ref{Heatmaps} visualizes the information in Tables \ref{FrenchAgree}-\ref{GermanAgree}, showing the heatmaps for head-layer combinations for average Agree-score for English, French, and German. In all cases, the agree information is spread over different heads, but is concentrated in a narrow field in the layers, with the highest scores being located in the early-mid layers.

\begin{table}[]
	\scalebox{.7}{
		\begin{tabular}{c|cccccccccccc}
			\hline 
			& Head=1 & Head=2 & Head=3 & Head=4 & Head=5 & Head=6 & Head=7 & Head=8 & Head=9 & Head=10 & Head=11 & Head=12 \\
			\hline 
			Layer=1  & 0.03   & 1.03   & 0.04   & 2.11   & 0.07   & 0.18   & 0.29   & 0.02   & 0.07   & 0.08    & 0.03    & 0.11    \\
			Layer=2  & 0.03   & 0.46   & 1.2    & 1.93   & \cellcolor{blue!30}4.98   & 0.05   & 0.09   & 0.21   & 0.04   & 0.05    & 0.05    & \cellcolor{blue!30}6.24    \\
			Layer=3  & 0.1    & 1.22   & 0.02   & \cellcolor{blue!30}6.88   & 0.59   & 0.06   & 0.04   & 0.12   & 0.05   & 0.05    & 0.08    & 0.13    \\
			Layer=4  & 0.06   & 0.05   & 0.08   & 0.04   & 0.44   & 0.1    & 0.07   & 0.06   & 0.03   & 0.37    & 0.15    & \cellcolor{blue!30}6.79    \\
			Layer=5  & 0.07   & 0.05   & 0.44   & 0.19   & 0.06   & 0.94   & 0.14   & 0.04   & 0.08   & 0.61    & 0.27    & 0.04    \\
			Layer=6  & 0.03   & \cellcolor{blue!30}6.35   & 0.17   & 0.09   & 0.19   & 0.4    & 0.31   & 0.16   & 1.14   & 0.7     & 0.24    & 0.04    \\
			Layer=7  & 0.05   & \cellcolor{blue!15}3.17   & 0.21   & 0.27   & 0.39   & 1.98   & 0.26   & 0.28   & 0.42   & \cellcolor{blue!30}4.97    & 0.19    & 0.05    \\
			Layer=8  & 2.61   & 0.17   & 0.31   & 0.45   & 0.5    & 0.06   & 0.21   & 0.04   & 0.17   & 0.16    & 0.08    & 0.08    \\
			Layer=9  & 0.15   & 0.1    & 0.05   & 0.19   & 0.3    & 0.05   & 0.18   & 0.1    & 0.72   & 0.58    & 0.1     & 0.07    \\
			Layer=10 & 0.09   & 0.08   & 0.72   & 0.23   & 0.87   & 0.06   & 0.07   & 0.09   & 0.14   & 0.18    & 0.08    & 0.1     \\
			Layer=11 & 0.07   & 0.14   & 0.06   & 0.06   & 0.08   & 0.06   & 0.12   & 0.15   & 0.16   & 0.15    & 0.05    & 0.08    \\
			Layer=12 & 0.05   & 0.06   & 0.05   & 0.09   & 0.04   & 0.05   & 0.04   & 0.06   & 0.03   & 0.04    & 0.06    & 0.07    \\
			\hline 
	\end{tabular}}
	\caption{Average agree-score (average $\chi^{2}$-score over distributions in agree-set) over English Agree dataset for each head/layer combination. Light blue shading denotes value exceeds $\chi^{2}$-value required for $p$-value$<0.1$ for one degree of freedom (2.706). Dark blue shading denotes value exceeds $\chi^{2}$-score for $p$-value$<0.05$ (3.841).} 
	\label{EnglishAgree} 
\end{table}

\begin{table}[h!]
	\scalebox{.7}{
		\begin{tabular}{c|cccccccccccc}
			\hline 
			& Head=1 & Head=2 & Head=3 & Head=4 & Head=5 & Head=6 & Head=7 & Head=8 & Head=9 & Head=10 & Head=11 & Head=12 \\
			\hline 
			Layer=1  & 0.02   & 0.17   & 0.03   & 0.03   & 0.01   & 0.43   & 0.04   & 0.04   & 0.03   & \cellcolor{blue!15}3.11    & 0.02    & 1.21    \\
			Layer=2  & 0.05   & 0.05   & 0.03   & 0.06   & 0.03   & 0.07   & \cellcolor{blue!15}3.39   & 0.03   & 0.06   & 0.02    & 0.08    & 0.67    \\
			Layer=3  & \cellcolor{blue!15}3.25   & 0.03   & 0.02   & \cellcolor{blue!30}4.42   & 0.58   & 0.56   & \cellcolor{blue!30}8.55   & 0.05   & 0.03   & 0.02    & 0.09    & 0.18    \\
			Layer=4  & \cellcolor{blue!30}8.55   & \cellcolor{blue!30}8.55   & \cellcolor{blue!30}8.61   & 0.21   & 0.25   & 1.87   & 1.15   & 0.02   & 0.05   & \cellcolor{blue!30}8.56    & 1.02    & 0.24    \\
			Layer=5  & 0.4    & 0.04   & \cellcolor{blue!15}3.84   & 0.12   & 0.37   & 1.49   & \cellcolor{blue!30}5.01   & 0.05   & \cellcolor{blue!30}5.92   & 0.05    & 0.77    & 0.03    \\
			Layer=6  & 0.06   & 2.05   & 0.7    & 1.36   & 0.02   & 1.91   & 0.76   & 0.07   & \cellcolor{blue!15}2.76   & 0.3     & 0.03    & \cellcolor{blue!15}3.71    \\
			Layer=7  & 0.99   & 0.09   & 0.49   & 0.03   & 0.02   & 0.03   & 0.03   & 0.81   & 0.14   & \cellcolor{blue!15}3.49    & 0.09    & 0.45    \\
			Layer=8  & 0.35   & 0.04   & 0.05   & 0.03   & 0.04   & 0.04   & 0.04   & 0.05   & 1.89   & 0.04    & 0.5     & 0.08    \\
			Layer=9  & 0.08   & 0.16   & 0.04   & 0.06   & 0.1    & 0.2    & 0.04   & 0.34   & 0.43   & 0.03    & 0.11    & 0.28    \\
			Layer=10 & 0.03   & 0.3    & 0.04   & 0.04   & 0.04   & 0.17   & 0.06   & 0.06   & 0.08   & 0.17    & 0.05    & 0.05    \\
			Layer=11 & 0.03   & 0.03   & 0.03   & 0.05   & 0.81   & 0.03   & 0.04   & 0.04   & 0.03   & 0.06    & 0.05    & 0.04    \\
			Layer=12 & 0.05   & 0.03   & 0.05   & 0.06   & 0.04   & 0.06   & 0.04   & 0.04   & 0.05   & 0.04    & 0.19    & 0.04    \\
			\hline 
	\end{tabular}}
	\caption{Average agree-score (average $\chi^{2}$-score over distributions in agree-set) over German Agree dataset for each head/layer combination. Light blue shading denotes value exceeds $\chi^{2}$-score required for $p$-value$<0.1$ for one degree of freedom (2.706). Dark blue shading denotes value exceeds $\chi^{2}$-score for $p$-value$<0.05$ (3.841).} 
	\label{GermanAgree} 
\end{table}

\section{Discussion}\label{Discussion}

Given the overall strong results on all languages for the classification task, it would appear that we can answer in the affirmative that BERT-style models are sensitive to morphological information at the featural level. Furthermore, this information appears to be encoded by models partitioning their space into convex sub-regions by feature-value, as feature values are largely recoverable by a linear classifier. Furthermore, supervision appears to aid significantly in extracting this morphological information, as initial attempts at unsupervised classification via a K-Means clustering task resulted in scores near the random baselines.

However, in spite of the success of the (supervised) classification tasks, there is room for improvement. Specifically, while syncretic forms are clearly not a problem for human language users, who effortlessly use context to parse the correct featural values from ambiguous forms, the same cannot be said for the models discussed here. The results in Table \ref{AmbiguityCorrelation} show ambiguity is negatively correlated with performance on classification, and to a significant degree in many cases. Thus, while the introduction of contextualized information into word embeddings no doubt helps to distinguish ambiguous forms (as shown in Figure \ref{GermanRussianCase}), BERT-style models have not solved the problem, at least not given the type of classifier examined in this work.

With regard to the location of morphological information in these models, it is typically assumed that in multi-layered models such as ELMo \cite{PetersEtAl2018a} and BERT, relatively ``shallow'' and local information is housed in the early layers, while information becomes more abstract and non-local as information progresses through the layers. This roughly follows a traditional NLP pipeline \cite{TenneyEtAl2019a}, and indeed standard linguistic thought on the process of the sound-to-meaning transduction. The results of the classification experiments shown here are therefore somewhat surprising at first glance, with layer being relatively uncorrelated with performance for English, French, and Spanish. One would expect morphological information of the kind necessary for feature-value classification to reside mostly in the middle-to-late layers.

However, the results for German and Russian show that each language in fact shows peak performance in the middle layers for overall performance (Figure \ref{Layers}), and the middle-to-late layers show the best ability to disambiguate syncretic forms (Figure \ref{GermanRussianCase}). This suggests that morphological information of the type considered here does reside mostly in these layers, in line with the findings of \newcite{PetersEtAl2018b} and \newcite{TenneyEtAl2019a}. We speculate that the exceptional performance of English, French, and Spanish throughout all layers is due to their simple morphological paradigms (relative to German and Russian), making morphological values more predictable from low-level information like orthography, and more fixed word-order syntax potentially making morphological information more recoverable from later layers.

The results from the Agree experiments likewise show that pre-trained BERT-style architectures are sensitive to morphological information. The agree information examined in these experiments is encoded in the attention coefficients of certain head-layer combinations which focus in on the agree relation, presumably passing morphological information between words in a sentence which stand in the relevant relation. This result further adds to the evidence that different head-layer combinations specialize for different types of linguistic information (see \newcite{ClarkEtAl2019}). Finally, the heatmaps in Figure \ref{Heatmaps} tentatively suggest that this information is best reflected in the early-mid layers.

\section{Conclusion}\label{Conclusion} 

This work has sought to address the question of the amount of morphological feature information in pre-trained BERT-style models for multiple Indo-European languages, using (i) classification tasks, and (ii) a task designed to identify the agree relation in attention distributions. The results show that the models examined are sensitive to morphological information of the type considered, with experiments showing strong performance for each language on the (supervised) feature-value classification tasks, and also that certain attention heads learn to focus attention in a manner consistent with morphological agreement. 

Furthermore, the findings here coincide with other work which suggests that morphological information may be best represented in the middle layers of deep contextualized language models like BERT. Given the results of this study, future work should include (i) identifying morphological information in BERT-style models in an unsupervised fashion, (ii) improving models' ability to disambiguate syncretic forms for languages with complex inflectional morphology, and (iii) further exploration of how morphological information is shared between words via self-attention.

\section*{Acknowledgements}

This paper has benefited from fruitful discussion with and helpful comments from John Goldsmith and Taeuk Kim.

\bibliographystyle{coling}
\bibliography{coling2020}

\appendix




\section{Treebank Details}\label{Treebanks}

For English, the following treebanks were used: the EWT treebank \cite{SilveiraEtAl14}, the GUM treebank \cite{Zeldes2017}, the LinES treebank \cite{Ahrenberg2007}, the English portion of ParTUT \cite{BoscoEtAl2012}, English-PUD \cite{ZemanEtAL2018}, and the English-Pronouns treebank \cite{Munro2020}. For French, the following treebanks were used: French Question Bank \cite{JudgeEtAl2006}, the GSD French treebank \cite{NivreEtAl2016}, French portion of ParTUT, French-PUD, Sequoia \cite{CanditoSeddah2012}, and the French Spoken Treebank, adapted from the Rhapsoide prosodic-syntactic treebank \cite{LacheretEtAl2014}. For German the following treebanks were used: The HDT-UD treebank \cite{VolkerEtAl2019}, and the GSD German treebank. For Russian, the following treebanks were used: The GSD Russian treebank, Russian-PUD, The SynTagRus treebank \cite{NivreEtAl2008}, and the Taiga treebank \cite{LyashevskayaEtAl2016}. For Spanish, the following treebanks were used: The AnCora treebank \cite{TauleEtAl2008}, the GSD Spanish treebank, and Spanish-PUD.

\section{Full Results}\label{FullResults}

Table \ref{ClassificationResultsByFeatureRandom} contains weighted F1 scores for each language feature combination where embeddings are from randomly initialized and untrained models, and serves as the baseline reference against which Table \ref{ClassificationResultsByFeature} should be considered.

\begin{table*}[h!]
	\small
	\begin{center}
		\begin{tabularx}{\linewidth}{X|X|X|X|X|X||X}
			\hline 
			Language  & English & French & German & Russian & Spanish & Average \\
			\hline
		\end{tabularx}
		\begin{tabularx}{\linewidth}{XXX|XXX|XXX|XXX|XXX|XXX||XXX}
			Feature$\backslash$Task &&& KM & Lin & NN & KM & Lin & NN & KM & Lin & NN & KM & Lin & NN & KM & Lin & NN & KM & Lin & NN \\
			\hline 
			
			Case     &&&  \cellcolor{lightgray}       &  \cellcolor{lightgray}       &   \cellcolor{lightgray}      & \cellcolor{lightgray}       &  \cellcolor{lightgray}      & \cellcolor{lightgray}       & 0.21   & 0.6    & 0.64   & 0.12    & 0.28    & 0.29    &    \cellcolor{lightgray}     &   \cellcolor{lightgray}      &   \cellcolor{lightgray}      & 0.16    & 0.44    & 0.47    \\
			
			Gender   &&&  \cellcolor{lightgray}       &  \cellcolor{lightgray}       &  \cellcolor{lightgray}       & 0.44   & 0.62   & 0.61   & 0.25   & 0.48   & 0.52   & 0.26    & 0.4     & 0.43    & 0.43    & 0.74    & 0.73    & 0.34    & 0.56    & 0.58    \\
			
			Mood     &&& 0.46    & 0.84    & 0.84    & 0.17   & 0.69   & 0.65   & 0.35   & 0.74   & 0.8    & 0.2     & 0.87    & 0.92    & 0.18    & 0.65    & 0.59    & 0.27    & 0.76    & 0.76    \\
			
			Number   &&& 0.39    & 0.63    & 0.61    & 0.39   & 0.7    & 0.62   & 0.4    & 0.58   & 0.55   & 0.38    & 0.52    & 0.53    & 0.34    & 0.66    & 0.66    & 0.38    & 0.62    & 0.6     \\
			
			Person   &&& 0.4     & 0.94    & 0.94    & 0.33   & 0.79   & 0.8    & 0.27   & 0.74   & 0.74   & 0.27    & 0.79    & 0.79    & 0.24    & 0.73    & 0.63    & 0.3     & 0.8     & 0.78    \\
			
			Tense    &&& 0.49    & 0.76    & 0.76    & 0.21   & 0.66   & 0.65   & 0.44   & 0.76   & 0.7    & 0.29    & 0.53    & 0.52    & 0.2     & 0.68    & 0.64    & 0.33    & 0.68    & 0.65    \\
			
			VerbForm &&& 0.2     & 0.51    & 0.56    & 0.22   & 0.57   & 0.56   & 0.27   & 0.58   & 0.53   & 0.2     & 0.46    & 0.45    & 0.2     & 0.47    & 0.47    & 0.22    & 0.52    & 0.51    \\
			
			\hline 
			\hline 
			
			Average  &&& 0.39    & 0.74    & 0.74    & 0.29   & 0.67   & 0.65   & 0.31   & 0.64   & 0.64   & 0.25    & 0.55    & 0.56    & 0.27    & 0.65    & 0.62    & 0.3     & 0.65    & 0.64    \\
			\hline 
		\end{tabularx}
		
	\end{center}
	\caption{Random baselines for weighted F1 Scores for each language and feature; scores averaged across all layers. Compare against Table \ref{ClassificationResultsByFeature}.}
	
	\label{ClassificationResultsByFeatureRandom}
\end{table*}

Table \ref{ClassificationResultsByLayer} houses the full results for the layer-wise classification scores. Table \ref{ClassificationResultsByLayerRandom} contains the random baselines against which to compare Table \ref{ClassificationResultsByLayer}, that is it reflects results from the same task, except the input embeddings were from randomly initialized, untrained models.

\begin{table*}[h!]
	\small
	\begin{center}
		\begin{tabularx}{\linewidth}{X|X|X|X|X|X||X}
			\hline 
			Language  & English & French & German & Russian & Spanish & Average \\
			\hline
		\end{tabularx}
		\begin{tabularx}{\linewidth}{XXX|XXX|XXX|XXX|XXX|XXX||XXX}
			Feature$\backslash$Task &&& KM & Lin & NN & KM & Lin & NN & KM & Lin & NN & KM & Lin & NN & KM & Lin & NN & KM & Lin & NN \\
			\hline 
			
			Layer=Input       &&& \textcolor{red}{0.32}    & 0.93    & 0.94    & 0.36   & 0.95   & 0.95   & 0.43   & 0.85   & 0.85   & \textcolor{red}{0.22}    & 0.89    & 0.9     & 0.29    & 0.94    & 0.94    & 0.33    & 0.91    & 0.92    \\
			
			Layer=1       &&& \textcolor{red}{0.38}    & 0.94    & 0.95    & \textbf{0.45}   & 0.96   & 0.96   & \textcolor{red}{0.31}   & 0.88   & 0.89   & 0.32    & 0.9     & 0.91    & \textcolor{red}{0.28}    & 0.96    & 0.95    & 0.35    & 0.93    & 0.93    \\
			
			Layer=2       &&& \textcolor{red}{0.24}    & 0.96    & 0.96    & 0.38   & 0.97   & 0.91   & 0.42   & 0.9    & 0.88   & \textcolor{red}{0.25}    & 0.91    & 0.92    & 0.34    & \textbf{0.97}    & \textbf{0.97}    & 0.33    & 0.94    & 0.93    \\
			
			Layer=3       &&& 0.48    & 0.96    & 0.96    & \textbf{0.45}   & 0.98   & 0.93   & 0.45   & 0.92   & 0.91   & 0.32    & 0.93    & 0.93    & 0.43    & \textbf{0.97}    & 0.96    & \textbf{0.43}    & 0.95    & 0.94    \\
			
			Layer=4       &&& \textcolor{red}{0.38}    & 0.96    & 0.96    & 0.43   & 0.98   & 0.98   & 0.34   & 0.94   & 0.93   & 0.32    & 0.94    & 0.94    & 0.3     & \textbf{0.97}    & 0.96    & 0.35    & 0.96    & \textbf{0.96}    \\
			
			Layer=5       &&& \textbf{0.6}     & 0.96    & 0.96    & 0.44   & 0.98   & 0.97   & 0.36   & 0.95   & \textbf{0.96}   & 0.3     & 0.95    & 0.95    & 0.33    & \textbf{0.97}    & 0.96    & 0.41    & 0.96    & \textbf{0.96}    \\
			
			Layer=6       &&& \textcolor{red}{0.31}    & \textbf{0.97}    & 0.96    & 0.34   & 0.98   & 0.97   & \textcolor{red}{0.3}    & \textbf{0.96}   & \textbf{0.96}   & \textcolor{red}{0.2}     & \textbf{0.96}    & \textbf{0.96}    & 0.34    & 0.96    & 0.92    & 0.3     & 0.96    & 0.95    \\
			
			Layer=7       &&& 0.49    & 0.96    & 0.96    & 0.42   & 0.98   & 0.98   & \textbf{0.51}   & \textbf{0.96}   & \textbf{0.96}   & \textcolor{red}{0.29}    & 0.95    & 0.95    & 0.37    & 0.96    & 0.96    & 0.42    & 0.96    & \textbf{0.96}    \\
			
			Layer=8       &&& 0.52    & \textbf{0.97}    & 0.96    & \textcolor{red}{0.3}    & 0.98   & 0.91   & 0.34   & 0.95   & \textbf{0.96}   & \textbf{0.36}    & \textbf{0.96}    & \textbf{0.96}    & 0.3     & 0.96    & 0.96    & 0.36    & 0.96    & 0.95    \\
			
			Layer=9       &&& 0.55    & \textbf{0.97}    & \textbf{0.97}    & 0.31   & \textbf{0.99}   & 0.98   & 0.39   & \textbf{0.96}   & \textbf{0.96}   & 0.25    & \textbf{0.96}    & \textbf{0.96}    & \textcolor{red}{0.28}    & 0.96    & 0.94    & 0.36    & \textbf{0.97}    & \textbf{0.96}    \\
			
			Layer=10      &&& 0.48    & 0.96    & 0.96    & \textcolor{red}{0.27}   & 0.98   & 0.98   & 0.48   & 0.95   & 0.95   & \textcolor{red}{0.22}    & \textbf{0.96}    & \textbf{0.96}    & \textbf{0.56}    & 0.96    & 0.95    & 0.4     & 0.96    & \textbf{0.96}    \\
			
			Layer=11      &&& 0.54    & 0.96    & 0.95    & 0.34   & \textbf{0.99}   & \textbf{0.99}   & \textcolor{red}{0.29}   & 0.94   & 0.93   & 0.3     & 0.95    & 0.95    & 0.34    & 0.95    & 0.95    & 0.36    & 0.96    & 0.95    \\
			
			Layer=12      &&& 0.45    & 0.96    & 0.92    & 0.38   & \textbf{0.99}   & 0.92   & 0.42   & 0.95   & 0.94   & 0.35    & 0.94    & 0.94    & 0.3     & 0.95    & 0.94    & 0.38    & 0.96    & 0.93    \\
			
			\hline\hline 
			
			Average &&& 0.44    & 0.96    & 0.96    & 0.38   & 0.98   & 0.96   & 0.39   & 0.93   & 0.93   & 0.28    & 0.94    & 0.94    & 0.34    & 0.96    & 0.95    & 0.37    & 0.95    & 0.95 \\
			\hline
		\end{tabularx}
		
	\end{center}
	\caption{F1 Scores by layer averaged across all relevant features. Bold indicates best score in column. Red indicates $\leq$ random baseline.}
	
	\label{ClassificationResultsByLayer}
\end{table*}

\begin{table*}[h!]
	\small
	\begin{center}
		\begin{tabularx}{\linewidth}{X|X|X|X|X|X||X}
			\hline 
			Language  & English & French & German & Russian & Spanish & Average \\
			\hline
		\end{tabularx}
		\begin{tabularx}{\linewidth}{XXX|XXX|XXX|XXX|XXX|XXX||XXX}
			Feature$\backslash$Task &&& KM & Lin & NN & KM & Lin & NN & KM & Lin & NN & KM & Lin & NN & KM & Lin & NN & KM & Lin & NN \\
			\hline 
			
			Layer=Input       &&& 0.42    & 0.77    & 0.78    & 0.28   & 0.72   & 0.69   & 0.3    & 0.68   & 0.69   & 0.23    & 0.6     & 0.61    & 0.25    & 0.7     & 0.71    & 0.3     & 0.69    & 0.7     \\
			
			Layer=1       &&& 0.45    & 0.76    & 0.78    & 0.31   & 0.71   & 0.69   & 0.34   & 0.68   & 0.67   & 0.25    & 0.59    & 0.59    & 0.29    & 0.7     & 0.67    & 0.33    & 0.69    & 0.68    \\
			
			Layer=2       &&& 0.43    & 0.75    & 0.77    & 0.33   & 0.71   & 0.67   & 0.32   & 0.67   & 0.66   & 0.29    & 0.58    & 0.6     & 0.22    & 0.68    & 0.66    & 0.32    & 0.68    & 0.67    \\
			
			Layer=3       &&& 0.43    & 0.75    & 0.77    & 0.3    & 0.7    & 0.71   & 0.29   & 0.65   & 0.65   & 0.25    & 0.57    & 0.55    & 0.23    & 0.68    & 0.69    & 0.3     & 0.67    & 0.67    \\
			
			Layer=4       &&& 0.42    & 0.76    & 0.78    & 0.26   & 0.68   & 0.68   & 0.31   & 0.65   & 0.67   & 0.24    & 0.56    & 0.57    & 0.23    & 0.68    & 0.55    & 0.29    & 0.67    & 0.65    \\
			
			Layer=5       &&& 0.34    & 0.74    & 0.75    & 0.36   & 0.68   & 0.67   & 0.31   & 0.65   & 0.67   & 0.22    & 0.55    & 0.55    & 0.26    & 0.66    & 0.67    & 0.3     & 0.66    & 0.66    \\
			
			Layer=6       &&& 0.35    & 0.74    & 0.74    & 0.24   & 0.68   & 0.67   & 0.3    & 0.63   & 0.63   & 0.22    & 0.55    & 0.58    & 0.29    & 0.65    & 0.58    & 0.28    & 0.65    & 0.64    \\
			
			Layer=7       &&& 0.36    & 0.73    & 0.72    & 0.32   & 0.67   & 0.65   & 0.34   & 0.63   & 0.67   & 0.3     & 0.54    & 0.57    & 0.26    & 0.65    & 0.58    & 0.31    & 0.64    & 0.64    \\
			
			Layer=8       &&& 0.38    & 0.73    & 0.74    & 0.3    & 0.65   & 0.64   & 0.33   & 0.64   & 0.61   & 0.22    & 0.54    & 0.55    & 0.28    & 0.62    & 0.62    & 0.3     & 0.64    & 0.63    \\
			
			Layer=9       &&& 0.35    & 0.72    & 0.74    & 0.29   & 0.64   & 0.65   & 0.32   & 0.62   & 0.62   & 0.24    & 0.52    & 0.56    & 0.28    & 0.62    & 0.53    & 0.3     & 0.62    & 0.62    \\
			
			Layer=10      &&& 0.38    & 0.72    & 0.72    & 0.3    & 0.63   & 0.62   & 0.28   & 0.62   & 0.65   & 0.25    & 0.51    & 0.53    & 0.28    & 0.62    & 0.5     & 0.3     & 0.62    & 0.6     \\
			
			Layer=11      &&& 0.35    & 0.7     & 0.73    & 0.26   & 0.63   & 0.58   & 0.31   & 0.61   & 0.56   & 0.23    & 0.5     & 0.54    & 0.3     & 0.61    & 0.53    & 0.29    & 0.61    & 0.59    \\
			
			Layer=12      &&& 0.38    & 0.71    & 0.68    & 0.26   & 0.62   & 0.63   & 0.35   & 0.6    & 0.62   & 0.25    & 0.51    & 0.49    & 0.28    & 0.61    & 0.57    & 0.3     & 0.61    & 0.6     \\
			
			\hline\hline 
			
			Average &&& 0.39    & 0.74    & 0.75    & 0.29   & 0.67   & 0.66   & 0.31   & 0.64   & 0.64   & 0.25    & 0.55    & 0.56    & 0.27    & 0.65    & 0.61    & 0.3     & 0.65    & 0.64   \\
			\hline
		\end{tabularx}
		
	\end{center}
	\caption{Random baselines for F1 Scores by layer averaged across all relevant features. Compare against Table \ref{ClassificationResultsByLayer}.}
	
	\label{ClassificationResultsByLayerRandom}
\end{table*}

Tables \ref{EnglishOut}-\ref{GermanOut} show English's, French's, and German's Out-score for each layer-head combination. 

\begin{table}[h!]
	\scalebox{.7}{
		\begin{tabular}{c|cccccccccccc}
			\hline 
			& Head=1 & Head=2 & Head=3 & Head=4 & Head=5 & Head=6 & Head=7 & Head=8 & Head=9 & Head=10 & Head=11 & Head=12 \\
			\hline 
			Layer=1  & 0.05   & 0.18   & 0.03   & 0.29   & 0.09   & 0.07   & 0.1    & 0.03   & 0.07   & 0.05    & 0.07    & 0.08    \\
			Layer=2  & 0.06   & 0.28   & 0.21   & 0.24   & 0.56   & 0.08   & 0.07   & 0.26   & 0.07   & 0.11    & 0.11    & 0.57    \\
			Layer=3  & 0.11   & 0.26   & 0.04   & 0.6    & 0.16   & 0.15   & 0.06   & 0.09   & 0.09   & 0.08    & 0.09    & 0.08    \\
			Layer=4  & 0.11   & 0.1    & 0.07   & 0.04   & 0.25   & 0.07   & 0.11   & 0.13   & 0.04   & 0.14    & 0.14    & 0.7     \\
			Layer=5  & 0.15   & 0.08   & 0.32   & 0.14   & 0.08   & 0.2    & 0.11   & 0.09   & 0.12   & 0.15    & 0.15    & 0.07    \\
			Layer=6  & 0.04   & 0.65   & 0.24   & 0.13   & 0.18   & 0.13   & 0.16   & 0.28   & 0.22   & 0.18    & 0.18    & 0.12    \\
			Layer=7  & 0.09   & 0.49   & 0.18   & 0.16   & 0.26   & 0.3    & 0.13   & 0.2    & 0.2    & 0.52    & 0.15    & 0.07    \\
			Layer=8  & 0.31   & 0.18   & 0.12   & 0.14   & 0.17   & 0.07   & 0.22   & 0.08   & 0.14   & 0.14    & 0.14    & 0.19    \\
			Layer=9  & 0.2    & 0.12   & 0.09   & 0.13   & 0.17   & 0.06   & 0.16   & 0.1    & 0.15   & 0.16    & 0.1     & 0.12    \\
			Layer=10 & 0.1    & 0.15   & 0.18   & 0.25   & 0.17   & 0.1    & 0.1    & 0.11   & 0.11   & 0.12    & 0.08    & 0.12    \\
			Layer=11 & 0.07   & 0.11   & 0.12   & 0.14   & 0.11   & 0.12   & 0.12   & 0.14   & 0.1    & 0.09    & 0.09    & 0.09    \\
			Layer=12 & 0.1    & 0.1    & 0.08   & 0.08   & 0.08   & 0.11   & 0.09   & 0.11   & 0.06   & 0.08    & 0.11    & 0.09       \\
			\hline 
	\end{tabular}}
	\caption{Average Out-score (average $\chi^{2}$-score over distributions not in agree-set) over English Agree dataset for each head/layer combination. } 
	\label{EnglishOut} 
\end{table}

\begin{table}[h!]
	\scalebox{.7}{
		\begin{tabular}{c|cccccccccccc}
			\hline 
			& Head=1 & Head=2 & Head=3 & Head=4 & Head=5 & Head=6 & Head=7 & Head=8 & Head=9 & Head=10 & Head=11 & Head=12 \\
			\hline 
			Layer=1  & 0.02   & 0.05   & 0.02   & 0.08   & 0.45   & 0.02   & 0.03   & 0.02   & 0.03   & 0.38    & 0.03    & 0.02    \\
			Layer=2  & 0.09   & 0.09   & 0.13   & 0.09   & 0.08   & 0.05   & 0.06   & 0.06   & 0.06   & 0.13    & 0.11    & 0.05    \\
			Layer=3  & 0.13   & 0.89   & 0.05   & 0.06   & 0.05   & 0.09   & 0.1    & 0.18   & 0.06   & 0.06    & 0.1     & 0.07    \\
			Layer=4  & 0.09   & 0.13   & 0.05   & 0.07   & 0.06   & 0.07   & 0.08   & 0.1    & 0.09   & 0.1     & 0.08    & 0.08    \\
			Layer=5  & 0.07   & 0.08   & 0.05   & 0.57   & 0.08   & 0.07   & 0.08   & 0.06   & 0.09   & 0.05    & 0.09    & 0.08    \\
			Layer=6  & 0.1    & 0.08   & 0.05   & 0.08   & 0.09   & 0.06   & 0.07   & 0.08   & 0.07   & 0.06    & 0.06    & 0.17    \\
			Layer=7  & 0.08   & 0.09   & 0.06   & 0.04   & 0.09   & 0.06   & 0.1    & 0.08   & 0.08   & 0.05    & 0.08    & 0.08    \\
			Layer=8  & 0.09   & 0.07   & 0.1    & 0.07   & 0.07   & 0.08   & 0.12   & 0.11   & 0.06   & 0.09    & 0.07    & 0.08    \\
			Layer=9  & 0.08   & 0.06   & 0.22   & 0.06   & 0.08   & 0.07   & 0.15   & 0.09   & 0.08   & 0.06    & 0.09    & 0.08    \\
			Layer=10 & 0.08   & 0.09   & 0.15   & 0.09   & 0.06   & 0.08   & 0.09   & 0.09   & 0.07   & 0.08    & 0.06    & 0.07    \\
			Layer=11 & 0.09   & 0.08   & 0.09   & 0.07   & 0.09   & 0.12   & 0.09   & 0.09   & 0.1    & 0.07    & 0.09    & 0.09    \\
			Layer=12 & 0.06   & 0.08   & 0.07   & 0.06   & 0.06   & 0.07   & 0.07   & 0.07   & 0.08   & 0.07    & 0.08    & 0.08      \\
			\hline 
	\end{tabular}}
	\caption{Average Out-score (average $\chi^{2}$-score over distributions not in agree-set) over French Agree dataset for each head/layer combination. Light blue shading denotes value exceeds $\chi^{2}$-value required for $p$-value$<0.1$ for one degree of freedom (2.706).} 
	\label{FrenchOut} 
\end{table}

\makeatletter
\setlength{\@fptop}{0pt}
\makeatother
\begin{table}[hbt!]
	\scalebox{.7}{
		\begin{tabular}{c|cccccccccccc}
			\hline 
			& Head=1 & Head=2 & Head=3 & Head=4 & Head=5 & Head=6 & Head=7 & Head=8 & Head=9 & Head=10 & Head=11 & Head=12 \\
			\hline 
			Layer=1  & 0.047  & 0.107  & 0.055  & 0.043  & 0.041  & 0.202  & 0.032  & 0.044  & 0.037  & 0.248   & 0.045   & 0.205   \\
			Layer=2  & 0.07   & 0.133  & 0.378  & 0.102  & 0.163  & 0.282  & 0.414  & 0.296  & 0.115  & 0.145   & 0.101   & 0.174   \\
			Layer=3  & 0.388  & 0.847  & 0.586  & 0.562  & 0.419  & 0.386  & 0.659  & 0.667  & 0.134  & 0.547   & 0.314   & 0.141   \\
			Layer=4  & 0.548  & 0.55   & 0.575  & 0.591  & 1.309  & 0.841  & 1.251  & 0.606  & 1.276  & 0.652   & 0.454   & 0.555   \\
			Layer=5  & 0.233  & 0.086  & 0.48   & 1.053  & 0.484  & 0.502  & 0.522  & 0.316  & 0.465  & 0.485   & 0.461   & 0.548   \\
			Layer=6  & 0.348  & 0.407  & 0.414  & 0.301  & 0.083  & 0.436  & 0.278  & 0.307  & 0.455  & 0.105   & 0.212   & 0.375   \\
			Layer=7  & 0.234  & 0.178  & 0.396  & 0.218  & 0.08   & 0.103  & 0.097  & 0.297  & 0.12   & 0.439   & 0.412   & 0.463   \\
			Layer=8  & 0.293  & 0.158  & 0.227  & 0.196  & 0.347  & 0.211  & 0.329  & 0.202  & 0.361  & 0.081   & 0.249   & 0.185   \\
			Layer=9  & 0.258  & 0.223  & 0.109  & 0.422  & 0.182  & 0.278  & 0.22   & 0.252  & 0.303  & 0.137   & 0.143   & 0.163   \\
			Layer=10 & 0.291  & 0.324  & 0.254  & 0.218  & 0.233  & 0.331  & 0.293  & 0.273  & 0.306  & 0.3     & 0.324   & 0.325   \\
			Layer=11 & 0.21   & 0.222  & 0.25   & 0.304  & 0.304  & 0.206  & 0.224  & 0.267  & 0.262  & 0.346   & 0.251   & 0.268   \\
			Layer=12 & 0.388  & 0.293  & 0.479  & 0.497  & 0.209  & 0.322  & 0.29   & 0.246  & 0.281  & 0.405   & 0.221   & 0.273  \\
			\hline 
	\end{tabular}}
	\caption{Average Out-score (average $\chi^{2}$-score over distributions not in agree-set) over German Agree dataset for each head/layer combination. } 
	\label{GermanOut} 
\end{table}

\end{document}